\documentclass[11pt,a4paper]{article}
\usepackage{acl2020}
\usepackage{times}
\usepackage{latexsym}

\usepackage{url}

\usepackage{amsmath}
\usepackage{amssymb}

\usepackage{booktabs} 
\usepackage{multirow}
\usepackage{enumitem}
\usepackage{hhline}
\usepackage{pgf}
\usepackage{lscape} 

\usepackage{array}
\newcolumntype{L}[1]{>{\raggedright\let\newline\\\arraybackslash\hspace{0pt}}p{#1}}
\newcolumntype{C}[1]{>{\centering\let\newline\\\arraybackslash\hspace{0pt}}m{#1}}
\newcolumntype{R}[1]{>{\raggedleft\let\newline\\\arraybackslash\hspace{0pt}}m{#1}}

\usepackage{graphicx}
\usepackage{multirow}
\usepackage{caption}
\usepackage{subcaption}
\usepackage[bottom]{footmisc} 
\usepackage{dblfloatfix}  
\usepackage{booktabs}
\usepackage[capposition=bottom]{floatrow} 

\aclfinalcopy 

\title{GiBERT: Introducing Linguistic Knowledge into BERT\\ through  a Lightweight Gated Injection Method}

\author{Nicole Peinelt$^{1,2}$  \and  Marek Rei$^{3}$ \and Maria Liakata$^{1,2,4}$\\
$^1$The Alan Turing Institute, UK \\
$^2$University of Warwick, UK \\
$^3$Imperial College London, UK \\
$^4$Queen Mary University of London, UK \\
  {\tt \{n.peinelt, m.liakata\}@warwick.ac.uk}, {\tt marek.rei@imperial.ac.uk}}

\date{}

\begin{document}
\maketitle
\begin{abstract}
Large pre-trained language models such as BERT have been the driving force behind recent improvements across many NLP tasks. 
However, BERT is only trained to predict missing words - either behind masks or in the next sentence - and has no knowledge of lexical, syntactic or semantic information beyond what it picks up through unsupervised pre-training. We propose a novel method to explicitly inject linguistic knowledge in the form of word embeddings into any layer of a pre-trained BERT. Our performance improvements on multiple semantic similarity datasets when injecting dependency-based and counter-fitted embeddings indicate that such information is beneficial and currently missing from the original model. Our qualitative analysis shows that counter-fitted embedding injection particularly helps with cases involving synonym pairs.
\end{abstract}

\section{Introduction}
\label{sec:intro}

With the recent success of pre-trained language models such as ELMo \citep{peters_deep_2018-1} and BERT~\citep{devlin_bert_2019} across many areas of NLP, there is increased interest in exploring how these  architectures can be further improved. One line of work aims at model compression, making BERT smaller and accessible while mostly preserving its performance \citep{xu_bert--theseus_2020,goyal_power-bert_2020,sanh_distilbert_2019,aguilar_knowledge_2020,lan_albert_2020,chen_adabert_2020}. Other studies seek to further enhance model performance by duplicating existing layers \citep{kao_further_2020} or introducing external information into BERT, such as information from knowledge bases \citep{peters_knowledge_2019,wang_k-adapter_2020} or multi-modal information \citep{lu_vilbert_2019,lin_interbert_2020}. 

Before the rise of contextualised models, transfer of pre-trained information between datasets and tasks in NLP was based on word embeddings.
Over multiple years, substantial effort was placed into the creation of such embeddings. While originally capturing mainly collocation patterns \citep{mikolov_efficient_2013,pennington_glove:_2014}, subsequent work enriched these embeddings with additional information, such as dependencies \citep{levy_dependency-based_2014}, subword information \citep{bojanowski_enriching_2017,luong_better_2013}, word prototypes \citep{huang_improving_2012} and semantic lexicons \citep{faruqui_retrofitting_2015}. As a result, there exists a wealth of pre-trained embedding resources for many languages in a unified format which could provide complementary information for contemporary pre-trained contextual models.

In this work, we propose a new method for injecting pre-trained embeddings into any layer of BERT's internal representation. 
Our approach differs from previous work by introducing linguistically-enriched embeddings directly into BERT through a novel injection method. We apply our method to multiple semantic similarity detection benchmark datasets and show that injecting pre-trained dependency-based and counter-fitted embeddings can further enhance BERT's performance. 
More  specifically, we make the following contributions:

\begin{enumerate}
\item We propose GiBERT - a lightweight gated method for injecting externally pre-trained embeddings into BERT (section \ref{sec:architecture}).
\item We provide ablation studies and detailed analysis for core model components (section \ref{sec:gating_ablation}).
\item We demonstrate that our model improves BERT's performance on multiple semantic similarity detection datasets.
Moreoever, when compared to multi-head attention injection, our gated injection method uses fewer parameters while achieving comparable performance for dependency embeddings and improved results for counter-fitted embeddings (section \ref{sec:complete_model}).
\item Our qualitative analysis
provides insights into GiBERT's improved performance, such as in cases of sentences pairs involving synonyms.  
(section \ref{sec:error_analysis}).
\end{enumerate}

\section{Related work}
\label{sec:background}

\paragraph{BERT modifications} Due to BERT's widespread success in NLP, many recent studies have focused on further improving BERT by introducing external information. Studies differ regarding the type of external information provided, the application area and their technical approach.
We broadly categorise existing approaches based on their modification method into input-related, external and internal. \emph{Input modifications} \citep{zhao_bert_2020,singh_constructing_2020,lai_simple_2020,ruan_fine-tuning_2020} adapt the information that is fed to BERT - e.g. feeding text triples separated by [SEP] tokens instead of sentence pairs as in \citet{lai_simple_2020} -  while leaving the architecture unchanged. \emph{Output modifications} \citep{xuan_fgn_2020,zhang_semantics-aware_2020} build on BERT's pre-trained representation by adding external information after the encoding step - e.g. combining it with additional semantic information as in \citet{zhang_semantics-aware_2020} - without changing BERT itself. By contrast,
\emph{internal modifications} introduce new information directly into BERT by adapting its internal architecture. Relatively few studies have taken this approach as this is technically more difficult and might increase the risk of so-called catastrophic forgetting - completely forgetting previous knowledge when learning new tasks \citep{french_catastrophic_1999,wen_few-shot_2018}. However, such modifications also offer the opportunity to directly harness BERT's powerful architecture to process the external information alongside the pretrained one. 
Most existing work on internal modifications has attempted to combine BERT's internal representation with visual and knowledge base information:
\citet{lu_vilbert_2019} modified BERT's transformer block with co-attention to integrate visual and textual information, while
\citet{lin_interbert_2020}  introduced a multimodal model which uses multi-head attention to integrate encoded image and text information between each transformer block.
\citet{peters_knowledge_2019} suggested a word-to-entity attention mechanism to incorporate external knowledge into BERT  and
\citet{wang_k-adapter_2020} proposed to inject factual and linguistic knowledge through separate adapter modules. 
Our approach differs from previous research as we propose to introduce external information with an addition-based mechanism which uses fewer parameters than existing attention-based techniques \citep{lu_vilbert_2019,lin_interbert_2020,peters_knowledge_2019}. We further incorporate a gating mechanism to scale injected information in an attempt to reduce the risk of catastrophic forgetting. Moreover, our work focuses on injecting pretrained word embeddings, rather than multimodal or knowledge base information as in previous studies.

\paragraph{Semantic similarity detection} Detecting paraphrases and semantically related posts in Community Question Answering requires modelling the semantic relationship between a text pair. This is a fundamental and well known NLP problem for which many methods have been proposed. Early work has focused on feature-engineering techniques, exploring various syntactic \citep{filice_kelp_2017}, semantic \citep{balchev_pmi-cool_2016} and lexical features \citep{tran_jaist_2015,almarwani_gw_qa_2017}.
Subsequent work attempted to model text pair relationships solely based on increasingly complex neural architectures \citep{deriu_swissalps_2017,wang_bilateral_2017,tan_multiway_2018} or by combining both approaches through hybrid techniques \citep{wu_ecnu_2017,feng_beihang-msra_2017,koreeda_bunji_2017}.
Most recently, contextual models such as ELMo \citep{peters_deep_2018-1} and BERT \citep{devlin_bert_2019} have reached state-of-the-art performance through pretraining large context-aware language models on vast amounts of textual data.
Our study joins up earlier lines of work with current state-of-the-art contextual representations by combining BERT with dependency-based and counter-fitted embeddings, previously shown to be useful for semantic similarity detection.

\section{Datasets and Tasks}
\label{sec:tasks}

We focus on the task of semantic similarity detection which is a fundamental problem in NLP and involves modelling the semantic relationship between two sentences in a binary classification setup.
We work with the following five widely used datasets which cover a range of related tasks and sizes (see Appendix \ref{appendix:datasets}).

\paragraph{MSRP}
The Microsoft Research Paraphrase dataset (MSRP) contains 5K pairs of sentences from news websites which were obtained based on heuristics and an SVM classifier. Gold labels are based on human binary annotations for sentential paraphrase detection \citep{dolan_automatically_2005}.

\paragraph{SemEval} The SemEval 2017 CQA dataset \citep{SemEval-2017:task3} consists of three subtasks involving posts from the online forum Qatar Living\footnote{https://www.qatarliving.com/}.
Each subtask provides an initial post as well as 10 posts which were retrieved by a search engine and annotated with binary labels by humans. The task requires the distinction between relevant and non-relevant posts. The original problem is a ranking setting, but since the gold labels are binary, we focus on a classification setup.
In \textbf{subtask A}, the posts are questions and comments from the same thread, in an answer relevancy detection scenario (26K instances). \textbf{Subtask B} is question paraphrase detection (4K instances).
\textbf{Subtask C} is similar to A but comments were retrieved from an external thread (47K).  We use the 2016 test set as the dev set and the 2017 test set as the test set.

\paragraph{Quora}
The Quora duplicate questions dataset 
is the largest of the selected datasets, consisting of more than 400k question pairs with binary labels.\footnote{https://engineering.quora.com/Semantic-Question-Matching-with-Deep-Learning} The task is to predict whether two questions are paraphrases, similar to SemEval subtask B. We use \citet{wang_bilateral_2017}'s train/dev/test set partition.\\

\noindent All of the above datasets provide two short texts, each usually a single sentence but sometimes consisting of multiple sentences. For simplicity, we refer to each short text as `sentence'. We frame the task as semantic similarity detection between two sentences in a binary classification task.

\section{GiBERT}
\label{sec:model}

\subsection{Architecture}
\label{sec:architecture}

\begin{figure*}[t]
  \centering
  \includegraphics[height=8.3cm]{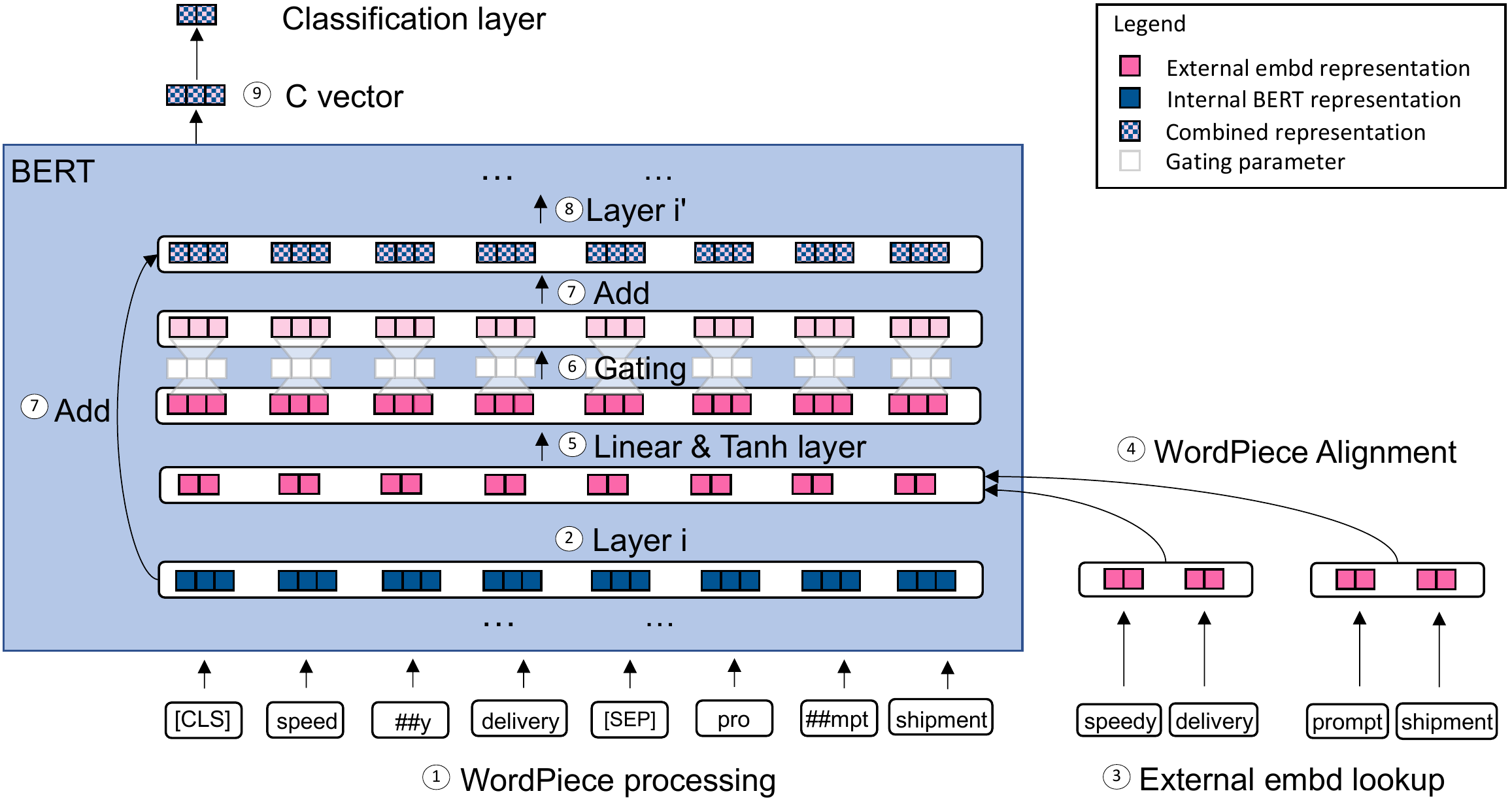}
   \caption{Our proposed GiBERT architecture illustrated with a toy example (where internal BERT dimension $d=3$ and embedding dimension $e=2$). The model input consists of a sentence pair which is processed with a WordPiece tokenizer (step 1) and encoded with BERT up to layer $i$ (step 2). We obtain an alternative representation for the sentence pair based on pretrained word embeddings (step 3), while ensuring that external word embeddings are aligned with BERT's WordPieces by repeating embeddings for tokens which have been broken down into several WordPieces (step 4). The aligned word embedding sequence is passed through a linear and tanh layer to match BERT's embedding dimension (step 5). We apply a gating mechanism (step 6) before adding the injected information to BERT's representation from layer i (step 7). The combined representation is passed on to the next layer (step 8). At the final layer, the C vector is used as the sentence pair representation, followed by a classification layer (step 9). 
} 
  \label{fig_model_GiBERT}
\end{figure*}

We propose  GiBERT - a \textbf{G}ated \textbf{I}njection Method for \textbf{BERT}.
GiBERT's architecture is illustrated with a toy example in Figure~\ref{fig_model_GiBERT} and comprises the following phases: obtaining BERT's intermediate representation from Transformer block $i$ (step 1-2 in Figure \ref{fig_model_GiBERT}), obtaining an alternative input representation based on linguistically-enriched word embeddings (step 3-4), combining both representations  (steps 5-7) and passing  on the injected information to subsequent BERT layers to make a final prediction (steps 8-9).

\paragraph{BERT representation}
We encode a sentence pair with a pre-trained BERT model (\citealt{devlin_bert_2019}) and obtain  BERT's internal representation at different layers (see section \ref{sec:injection_location} for injection layer choices).\footnote{We use the the uncased version of BERT$_\text{BASE}$ available through Tensorflow Hub.}
Following standard practice, we process the two input sentences $S_1$ and $S_2$ with a word piece tokenizer \citep{wu_googles_2016} and combine them using `[CLS]' and `[SEP]' tokens which indicate sentence boundaries. Then, the word pieces are mapped  to ids, resulting in a sequence of word piece ids $\mathbf{E^W} = [w_1,...,w_N]$ where $N$ indicates the number of word pieces in the sequence (step 1 in Figure \ref{fig_model_GiBERT}). 
 In the case of embedding layer injection, we use BERT's  embedding layer output denoted with $\mathbf{H^{0}}$ which results from summing the word piece embeddings $\mathbf{E^{W}}$, positional embeddings $\mathbf{E^{P}}$ and segment embeddings $\mathbf{E^{S}}$ (step 2):
\begin{align}
\label{eq_embedding_block}
\begin{split}
\mathbf{H^{0}} = \text{LayerNorm(}\mathbf{E^{W}}+\mathbf{E^{P}}+\mathbf{E^{S}})\\
\mathbf{E^{W}},\mathbf{E^{P}},\mathbf{E^{S}}, \mathbf{H^{0}} \in \mathbb{R}^{N \times D}
\end{split}
\end{align}
where $D$ is the internal hidden size of BERT ($D=768$ for BERT$_\text{BASE}$).
For injecting information at later layers, we obtain BERT's internal  representation $\mathbf{H^{i}} \in \mathbb{R}^{N \times D}$ after transformer block $i$ with $1 \leq i \leq L$ (step 2):
\begin{align}
\label{eq_transformer_block}
\begin{split}
\mathbf{M^{i}} &\text{= LayerNorm(}\mathbf{H^{i-1}}\text{+ MultiheadAtt(}\mathbf{H^{i-1}})) \\
\mathbf{H^{i}} &\text{= LayerNorm(}\mathbf{M^{i}}\text{+ FeedForward(}\mathbf{M^{i}}))
\end{split}
\end{align}
where L is the number of Transformer blocks ($L=12$ for BERT$_\text{BASE}$) and MultiheadAtt denotes multihead attention.

\paragraph{External embedding representation}
To enrich this representation, we obtain alternative representations for $S_1$ and $S_2$ by looking up word embeddings in a matrix of pre-trained embeddings $\mathbf{E} \in \mathbb{R}^{|V| \times E}$ where $|V|$ indicates the vocabulary size and $E$  is the dimensionality of the pre-trained embeddings (step 3, refer to section \ref{sec:injected_embd} for details on our choice of pre-trained embeddings).
To ensure alignment between BERT's representation at word piece level and the word embedding representation 
at token level, an alignment function copies embeddings of tokens that were separated into multiple word pieces and adds BERT's special `[CLS]' and `[SEP]' tokens, resulting in an injection sequence $\mathbf{I} \in \mathbb{R}^{P \times E}$ (step 4). For example, we copy the pre-trained embedding of the word `prompt' to match the two corresponding word pieces `pro' and `\#\#mpt' (see Figure \ref{fig_model_GiBERT}).

\paragraph{Multihead Attention Injection}
 Multihead attention was proposed by  \citet{vaswani_attention_2017}:
\begin{align}
\label{eq_multiheadattention_injection_1}
\begin{split}
&\text{MultiheadAtt}(\mathbf{Q},\mathbf{K},\mathbf{V}) \text{=} [\text{head}_1;...;\text{head}_h] \mathbf{W^O}\\
&\text{where } \text{head}_i \text{=} \text{Attention}(\mathbf{Q}\mathbf{W_i^Q},\mathbf{K}\mathbf{W_i^K},\mathbf{V}\mathbf{W_i^V})\\
\end{split}
\end{align}
and is employed in Transformer networks in the form of self-attention (where queries $\mathbf{Q}$, keys $\mathbf{K}$ and values $\mathbf{V}$ come from the previous layer) or encoder-decoder attention (where queries come from the decoder; keys and values from the encoder).
Previous work has successfully employed multihead attention to combine BERT with external information (see section \ref{sec:background}).
For example, in their multimodal VilBERT model, \citet{lu_vilbert_2019}  combined textual and visual representations by passing the keys and values from each modality as input to the other modality’s multi-head attention block.
Similarly, \citet{peters_knowledge_2019} used multihead attention to combine projected BERT representations (as queries) with entity-span representations (as keys and values) n their knowledge-enrichment method for BERT. 
For our case of combining BERT with the injection sequence, it is therefore intuitive to try to use the following multi-head attention injection method:
\begin{align}
\label{eq_multiheadattention_injection_2}
\begin{split}
\mathbf{H^{i'}} = \mathbf{H^{i}}+\text{MultiHeadAtt(}\mathbf{H^{i}}, \mathbf{I},  \mathbf{I})
\end{split}
\end{align}
where queries are provided by BERT's internal representation, while keys and values come from the injected embeddings.
The output of the attention mechanism is then combined with the previous layer through addition. 

\paragraph{Gated Injection} The above multihead attention injection mechanism is rather complex and requires many parameters. We therefore propose an alternative way of combining external embeddings with BERT which requires only 14\% of parameters used in multi-head attention (see Appendix \ref{appendix:parameters}).
First, we add a feed-forward layer -- consisting of a linear layer with $\mathbf{W^P} \in \mathbb{R}^{D \times E}$ and $\mathbf{b^P} \in \mathbb{R}^{D}$ with a tanh activation function -- to project the aligned embedding sequence to BERT's internal dimensions and squash the output values to a range between -1 and 1 (step 5):
\begin{align}
\label{eq_gating_projection}
\mathbf{P} = \text{FeedForward(}\mathbf{I}) \in \mathbf{R}^{N \times D}
\end{align}
Then, we use a residual connection  to inject the projected external information into BERT's representation from Transformer block $i$ (see section \ref{sec:injection_location} for injection at different locations) and obtain a new enriched representation $\mathbf{H^{i'}} \in \mathbb{R}^{N \times D}$:
\begin{align}
\label{eq_not_gating}
\mathbf{H^{i'}} = \mathbf{H^{i}} + \mathbf{P} 
\end{align}
However, as injection values can get rather large (between -1 and 1) in comparison to BERT's internal representation (based on our observation usually ranging around -0.1 to 0.1),
a downside of directly injecting external information in this way is that BERT's pre-trained information can be easily overwritten by the injection, resulting in catastrophic forgetting. 
To address this potential pitfall, we further propose a gating mechanism which uses a gating vector $\mathbf{g} \in \mathbb{R}^{D}$ to scale the injected information before combining it with BERT's internal representation as follows:
\begin{align}
\label{eq_gating}
\mathbf{H^{i'}} = \mathbf{H^{i}} + \mathbf{g} \odot \mathbf{P}
\end{align}
where $\odot$ denotes element-wise multiplication using broadcasting (step 6 \& 7).
The gating parameters are initialised with zeros and updated during training.
This has the benefit of starting finetuning from  representations which are equivalent to vanilla BERT and gradually introducing the injected information during finetuning along certain dimensions. In case the external representations are not beneficial for the task, it is easy for the model to ignore them by keeping the gating parameters at zero.

\paragraph{Output layer}
The combined representation $\mathbf{H^{i'}}$ is then fed as input to BERT's next Transformer block $i+1$ (step 8). At the final Transformer block $\mathbf{L}$, we use the $\mathbf{c} \in \mathbb{R}^{D}$ vector which corresponds to the `[CLS]' token in the input and is typically used as the sentence pair representation (step 9).
As proposed by  \citet{devlin_bert_2019}, this is followed by a softmax classification layer (with weights $\mathbf{W^L} \in \mathbb{R}^{C \times D}$ and $\mathbf{b^L} \in \mathbb{R}^{C}$)  to calculate class probablilities where $C$ indicates the number of classes.
During finetuning, we train the entire model for 3 epochs with early stopping and cross-entropy loss. 
Learning rates are tuned for each seed and dataset based on development set performance
(reported in Appendix \ref{appendix:hyperparams}).

\subsection{Injected Embeddings}
\label{sec:injected_embd}

While any kind of information could be injected, we focus on two types of pretrained embeddings: dependency-based \citep{levy_dependency-based_2014} and counter-fitted embeddings \citep{mrksic_counter-fitting_2016}. Our choice is motivated by previous research which found syntactic features useful for semantic similarity detection \citep{filice_kelp_2017,feng_beihang-msra_2017} and
counter-fitted embeddings helpful in several other tasks \citep{alzantot_generating_2018,jin_is_2020}.

The dependency-based embeddings by \citet{levy_dependency-based_2014} extend the SkipGram embedding algorithm proposed by \citet{mikolov_efficient_2013} by replacing linear bag-of-word contexts with dependency-based contexts which are extracted from parsed English Wikipedia sentences.
As BERT has not been exposed to dependencies during pretraining and previous studies have found that BERT's knowledge of syntax is only partial \citep{rogers_primer_2020}, we reason that these embeddings could provide helpful complementary information. 

The counter-fitted embeddings by \citet{mrksic_counter-fitting_2016}  integrate antonymy and synonymy relations into word embeddings based on an objective function which combines three principles: repelling antonyms, attracting synonymy  and preserving the vector space. For training, they obtain synoynmy and antonymy pairs from the Paraphrase Database and WordNet, demonstrating an increased performance on SimLex-999. We use their highest-scoring vectors which they obtained by applying their counter-fitting method to Paragram vectors from \citet{wieting_paraphrase_2015}. We reason that the antonym and synonym relations contained in the word embeddings could be especially useful for paraphrase detection by explicitly capturing these semantic relations.

\subsection{Injection Settings}

\begin{table}[t!]
  \small
  \centering
    \begin{tabular}{lR{0.9cm}R{0.795cm}R{0.5536cm}R{0.5536cm}R{0.5536cm}}
    \toprule
{} &  MSRP &  Quora &  \multicolumn{3}{c}{SemEval}  \\ 
  &  &  &  A & B & C \\
    	\midrule
BERT & .906 & .906 & .714 & .754 & .414 \\
\midrule
\multicolumn{6}{l}{GiBERT with dependency embeddings}\\
- no gating  & .906  &  .905  &   .732   &  .751  &    .424   \\ 
- with gating & \bf{.913} & \bf{.908}  & \bf{.755}   &  \bf{.778}  &   \bf{.433}   \\ 
\midrule
\multicolumn{6}{l}{GiBERT with counter-fitted embeddings}\\
- no gating  & \bf{.907}  &  .906  &   .733   &  .763  &  .435  \\ 
- with gating   & \bf{.907} &  \bf{.908} &  \bf{.751}   &  \bf{.767}   &  \bf{.451}    \\ 
    \bottomrule
    \end{tabular}%
  \caption{
  F1 development scores of models injecting pretrained embeddings after the embedding layer with vs. without gating mechanism. 
  }
       \label{tab:scaling}
\end{table}

\paragraph{Gating Mechanism}
\label{sec:gating_ablation}

Catastrophic forgetting is a potential problem when introducing external information into a pre-trained model as the injected information could disturb or completely overwrite existing knowledge \citep{wang_k-adapter_2020}.
In our proposed model, a gating mechanism is used to scale injected embeddings before adding them to the pre-trained internal BERT representation (see section \ref{sec:architecture}). To understand the importance of this mechanism, we contrast development set performance for injecting information after the embedding layer with gating - as defined in equation \ref{eq_gating} - and without - as in equation \ref{eq_not_gating} - (Table \ref{tab:scaling}). For dependency embedding injection without gating, performance only improves on 2 out of 5 datasets over the baseline and in some cases even drops below BERT's performance, while it outperforms the baseline on all datasets when using the gating mechanism.
Counter-fitted embedding injection without gating improves on 4 out of 5 datasets, with further improvements when adding gating, outperforming the vanilla BERT model across all datasets. In addition, gating makes model training more stable and reduces failed runs (where the model predicted only the majority class) from 30\% to 0\% on the particularly imbalanced SemEval C dataset. This highlights the importance of the gating mechanism in our proposed architecture. 

\paragraph{Injection Location}
\label{sec:injection_location}

In our proposed model, information can be injected between any of BERT's pre-trained transformer blocks. We reason that different locations may be more appropriate for certain kinds of embeddings as previous research has found that different types of information tend to be encoded and processed at specific BERT layers \citep{rogers_primer_2020}. We experiment with injecting embeddings at three possible locations: after the embedding layer (using $\mathbf{H^0}$), after the middle layer (using $\mathbf{H^6}$ in BERT$_\text{BASE}$) and after the penultimate layer (using $\mathbf{H^{11}}$ in BERT$_\text{BASE}$). 
Table \ref{tab:injection_layers} shows that  midlayer injection is ideal for counter-fitted embeddings, while late injection appears to work best for dependency embeddings (Table \ref{tab:injection_layers}).
This is in line with previous work which found that BERT tends to processes syntactic information at later layers than linear word-level information \citep{rogers_primer_2020}.
We consequently use these injection locations in our final model.

\begin{table}[t!]
  \small
    \begin{tabular}{lR{0.9cm}R{0.795cm}R{0.5536cm}R{0.5536cm}R{0.5536cm}}
    \toprule
{} &  MSRP &  Quora &  \multicolumn{3}{c}{SemEval}  \\ 
  &  &  &  A & B & C \\
    	\midrule
BERT & .906 & .906 & .714 & .754 & .414 \\
\midrule
\multicolumn{6}{l}{GiBERT with dependency embeddings}\\
- embd layer & .913 & .908  & .755   &  \bf{.778}  &   .433    \\ 
- layer 6  & .911  &  .908  &  .755   &  .776 &   .438   \\ 
- layer 11  & \bf{.914} &  \bf{.910} &  \bf{.760}  &  .773  &    \bf{.444}  \\ 
\midrule
\multicolumn{6}{l}{GiBERT with counter-fitted embeddings}\\
- embd layer & .907 &  .908 &  .751   &  .767   &  .451    \\ 
- layer 6   & \bf{.917} & \bf{.909}  &  \bf{.760}   &  \bf{.771}   &  \bf{.464}  \\ 
- layer 11   & .910 & .907  &  .755   &  \bf{.771}   &  .450  \\ 
    \bottomrule
    \end{tabular}%
   \caption{
  F1 scores of embedding injection at different layers on the development set. 
   }
       \label{tab:injection_layers}
\end{table}

\begin{table*}[b!]
  \small
  \centering
\begin{tabular}{lR{0.9cm}R{0.795cm}R{0.5536cm}R{0.5536cm}R{0.5536cm}|R{0.9cm}R{0.795cm}R{0.5536cm}R{0.5536cm}R{0.5536cm}}
\toprule
  & \multicolumn{5}{c}{F1} & \multicolumn{5}{c}{non-obvious F1} \\
{} &  MSRP &  Quora &  \multicolumn{3}{c}{SemEval} &  MSRP &  Quora &  \multicolumn{3}{c}{SemEval}\\
  &  &  &  A & B & C & &  &  A & B & C  \\
\midrule
\multicolumn{11}{l}{\bf{Previous systems}}\\
\newcite{filice_kelp_2017}  & - & - & - & .506 & - & - &- &- & .199 & - \\ 
\newcite{wu_ecnu_2017} & - & - & .777 & - & - & - & - & .707 & - & - \\
\newcite{koreeda_bunji_2017}   & - & - & - & - & .197 & - & - & - & - & .028 \\
\newcite{pang_text_2016}   & .829 & - & - & - & - & - & - & - & - &- \\ 
\newcite{gong_natural_2018} (accuracy)   & - & (.891) & - & - & - & - & - & - & - &- \\ 
 \newcite{zhang_semantics-aware_2020}* & .882 & .718 & - & - & - & - & - & - & - &- \\ 
\midrule
\multicolumn{11}{l}{\bf{Our implementation}}\\
BERT               &  .876 &   .902 &       .704 &       .473 &       .268 &  .827 &   .860 &       .656 &       .243 &        .085 \\
AiBERT$_{\text{dependency}}$    &  .871	&	.903	&.745	&.495	& \bf{.272}&	.827	&	.866&	.680&	.248	& \bf{.092} \\
GiBERT$_{\text{dependency}}$ & .883 & .904  & .768 & .474 & .238 & .849 & .864 & .704 & .231 & .087\\
AiBERT$_{\text{counter-fitted}}$ &  .877	& 	.904 &	.724 &	.496	& .263	& .835	 & \bf{.867} &	.662 &	\bf{.264} &	.076\\
GiBERT$_{\text{counter-fitted}}$ & \bf{.884} & \bf{.907} & \bf{.780} & \bf{.511} & .256 & \bf{.858} & .862 & \bf{.719} & .248 & .090 \\
\bottomrule
    \end{tabular}%
  \caption{Model performance on test set. All BERT-based methods use BERT$_{\text{BASE}}$.
 The first 6 rows are taken from the cited papers, the rest are our implementations.  Bold font highlights the best system. * indicates that the system reported performance on a slightly different dataset version. 
  }
      \label{tab:results}
\end{table*}%

\section{Evaluation}
\label{sec:evaluation}

\paragraph{Metrics} Our main evaluation metric is F1 score as this is more meaningful for datasets with imbalanced label distributions (such as SemEval C) than accuracy. We also report performance on difficult cases using the non-obvious F1 score \citep{peinelt_aiming_2019}. This metric distinguishes non-obvious instances in a dataset from obvious ones based on lexical overlap and gold labels, calculating a separate F1 score for challenging cases. It therefore tends to be lower than the normal F1 score.

\paragraph{Tuning}
\citet{dodge_fine-tuning_2020} recently showed that early stopping and random seeds can have considerable impact on the performance of finetuned BERT models. In this work, we finetune all models for 3 epochs with early stopping. Our reported scores average model performance across two different seeds for BERT-based models.

\subsection{Baselines}
\label{sec:baseline}

\paragraph{BERT}  
Following standard practice, we encode the sentence pair with BERT's $C$ vector from the final layer,
followed by a softmax layer. We finetune all layers for 3 epochs with early stopping. Following \citet{devlin_bert_2019}, we tune learning rates on the dev set of each dataset.

\paragraph{AiBERT}
We further provide an alternative \textbf{A}ttention-based embedding \textbf{I}njection method for \textbf{BERT} based on the multihead attention injection mechanism described in equations \ref{eq_multiheadattention_injection_1} to \ref{eq_multiheadattention_injection_2}. 
For direct comparison, we inject embeddings at the same layers as GiBERT (layer 6 for counter-fitted embeddings and layer 11 for dependency-based embeddings).
We follow the same finetuning procedure as GiBERT and the BERT baseline.

\paragraph{Previous systems} 
For SemEval, we compare against the best participating SemEval 2017 system for each subtask based on F1 score.
For MSRP, we show a neural matching architecture \citep{pang_text_2016}. 
For Quora, we compare against the Interactive Inference Network \citep{gong_natural_2018} using accuracy, as no F1 has been reported.
We also provide a semantics-aware BERT  model \citep{zhang_semantics-aware_2020} which leverages a semantic role labeler.

\begin{table*}[t!]
  \small
  \centering
    \begin{tabular}{L{0.04cm}L{5.5176cm}L{4.87cm}L{0.635cm}L{1.28264cm}L{1.28264cm}}
    \toprule
& Sentence 1 &  Sentence 2 &  Gold label &  BERT prediction & GiBERT prediction \\
    	\midrule
(1) & it took me more than 10 people; over the course of the whole day to convince my point at qatar \textbf{airways}... as to how my points needs to be redeemed... at long last my point was made... 
dont seem know what they are doing??? 
appalling to say the least &
this isn't the first time. so many rants by irate customers on so many diverse situations signals a very serious problem. so called first class \textbf{airlines} and no basic customer care. over confidence much? & is related  & not related  & is related \\
\midrule
(2) & hi; my \textbf{wife} was on a visit visa; today; her residency visa was issued; so i went to immigration and paid 500 so there is no need to leave the country and enter again on the residency visa . she has done her medical before for the visit visa extension; do we need to do the medical again for the residency visa? thanks
& dear all; please let me know how many days taking for approve family visa nw; am last wednesday (12/09/2012) apply family visa for my \textbf{husband} and daughter; but still now showing in moi website itz under review; itz usual reply? why delayed like this? please help me regards divya & is related  & is related  & not related \\
    \bottomrule
    \end{tabular}%
   \caption{Examples from the Semeval development set. Synonym and antonym pairs are highlighted in bold.}
       \label{tab:example}
\end{table*}%

\section{Results}
\label{sec:results}

\paragraph{Comparison with previous systems}
\label{sec:complete_model}
GiBERT with counter-fitted embeddings outperforms the F1 score of BERT and other previous systems across all datasets (except on SemEval~C)\footnote{As reflected by the lower scores compared to other datasets, SemEval C is particularly difficult due to the external question-answering scenario and its highly imbalanced label distribution which varies between train, dev and test set.}, see Table \ref{tab:results}.  It also improves the performance on non-obvious cases in comparison to previous systems.
The largest improvement of GiBERT is observed with counter-fitted embeddings, especially on the internal CQA datasets SemEval A and B (the datasets with the highest proportion of examples involving synonym pairs, see section \ref{sec:error_analysis}). 
GiBERT with dependency embeddings still generally improves over vanilla BERT, 
but performance gains tend to be smaller than with counter-fitted embeddings, possibly because semantic information tends to be more important for the tasks at hand.

\paragraph{Injection method} 
When contrasting the gated injection method (GiBERT) with an alternative attention-based injection method (AiBERT),
we find that both injection methods generally improve over the performance of the BERT baseline.
In direct comparison between both methods, we find that injecting embeddings with the lightweight gated method achieves comparable results to the complex multihead attention injection method for introducing dependency embeddings, while for the injection of counter-fitted embeddings, GiBERT even outperforms AiBERT.

\begin{table}[b!]
  \small
    \begin{tabular}{L{1.2cm}rrrrr}
    \toprule
{} &  MSRP &  Quora &  \multicolumn{3}{c}{SemEval}  \\ 
  &  &  &  A & B & C \\
\midrule
\multicolumn{6}{c}{Instances with \textbf{antonym} pairs}\\
 & (4\%)  & (4\%) &   (21\%) &   (28\%) &   (20\%)  \\ 
BERT & \bf{.81} & \bf{.87}  &  \bf{.77}  & \bf{.75}  &  \bf{.46}    \\ 
GiBERT  & \bf{.81} & .86  & \bf{.77}   & \bf{.75}  &  \bf{.46}   \\ 
\midrule
\multicolumn{6}{c}{Instances with \textbf{synonym} pairs}\\
  & (11\%)  &  (9\%) &   (22\%)  &  (31\%) &   (17\%) \\ 
BERT & .87 & .90&  .81  & .78  &   \bf{.54}   \\ 
GiBERT  & \bf{.90} & \bf{.91}  &  \bf{.82}  & \bf{.83}  &   \bf{.54}  \\ 
\midrule
\multicolumn{6}{c}{Instances \textbf{without} synonym/antonym pairs}\\
  & (85\%) &  (87\%)  &   (64\%) &   (51\%) &   (68\%)  \\ 
BERT & .91 & \bf{.91}  &  .71  & .72  &    .36  \\ 
GiBERT &  \bf{.92} & \bf{.91}  &  \bf{.73}  & \bf{.73}  & \bf{.41}    \\ 
    \bottomrule
    \end{tabular}%
   \caption{
F1 score on instances containing synonymy pairs, antonymy pairs or no pairs across datasets. (The added percentage of the three groups can exceed 100 as an instance can contain synonym and antonym pairs.)
   }
       \label{tab:synonym_antonym}
\end{table}


\paragraph{Error Analysis}
\label{sec:error_analysis}
Counter-fitted embeddings are designed to explicitly encode synonym and antonym relationships between words.
To better understand how the injection of counter-fitted embeddings affects the ability of our model to deal with instances involving such semantic relations, we use synonym and antonym pairs from the PPDB and wordnet (provided by \citealt{mrksic_counter-fitting_2016}) and search the development partition of the datasets for sentence pairs where the first sentence contains one word of the synonym/antonym pair and the second sentence the other word. 
 Table \ref{tab:synonym_antonym} reports F1 performance of our model on cases with synonym pairs, antonym pairs and neither one. 
We find that our model's F1 performance particularly improves over BERT on instances containing synonym pairs, as illustrated in example (1) in Table \ref{tab:example}.
By contrast, the performance on cases with antonym pairs stays roughly the same, even slightly decreasing on Quora. This can be understood with the help of example (2) in Table \ref{tab:example}, as word pairs can be antonyms in isolation (e.g. husband - wife), but not in the specific context of a given example (e.g. it's not important if the visa is for the wife or husband). In rare cases, the injection of distant antonym pair embeddings could deter the model from detecting related sentence pairs.
We also observe a slight performance boost for cases that don't contain synonym or antonym pairs. This could be because of improved representations for  words which occurred in examples without their synonym or antonym counterpart.

\section{Conclusion}
\label{sec:conclusion}

In this paper, we introduced a new approach for injecting external information into BERT. Our proposed method adds  linguistically enriched embeddings to BERT's internal representation through a lightweight gating mechanism which requires significantly fewer parameters than a multihead attention injection method. 
Evaluating our injection method on  multiple semantic similarity detection datasets, we demonstrated that injecting counter-fitted embeddings clearly improved performance over vanilla BERT, while dependency embeddings achieved slightly smaller gains for these tasks. 
In comparison to the multihead attention injection mechanism, we found the gated method at least as effective, with comparable performance for dependency embedding and improved results for counter-fitted embeddings.
Our qualitative analysis highlighted that counter-fitted injection was particularly helpful for instances involving synonym pairs.
Future work could explore combining multiple embedding sources or injecting other types of information. 
Another direction is to investigate the usefulness of embedding injection for other tasks or compressed BERT models. 

\bibliography{main.bib}
\bibliographystyle{acl_natbib}

\newpage
\onecolumn

\appendix

\section*{Appendix}
\label{sec:appendix}

\section{Datasets}
\label{appendix:datasets}

\begin{table*}[h!]
  \small
  \centering
    \begin{tabular}{L{1.3cm}L{3.5cm}L{3.5cm}L{3.6cm}l}
    \toprule
\multicolumn{1}{c}{Dataset}  & \multicolumn{1}{c}{Task}  & \multicolumn{3}{c}{Example}\\
& &  Sentence 1 & Sentence 2 & Label \\ 
    	\midrule
Quora  & Paraphrase detection  & There are only 2,000 Roman Catholics living in Banja Luka now. & There are just a handful of Catholics
left in Banja Luka. &  is\_paraphrase \\
MSRP & Paraphrase detection  &Which is the best way to learn coding? & How do you learn to program? & is\_paraphrase \\
SemEval  & (A) Internal answer detection   & Anybody recommend a good dentist in Doha? &Dr Sarah Dental Clinic & is\_related   \\ 
  & (B) Paraphrase detection   & Where I can buy good oil for massage? &Blackheads - Any suggestions on how 0 to get rid of them??&  not\_related \\
 & (C) External answer detection  & Can anybody tell me where is Doha clinic? &  Dr. Rizwi - Al Ahli Hospital& not\_related \\
    \bottomrule
    \end{tabular}%
   \caption{Text pair similarity data sets with examples.}
       \label{tab:cQA_datasets}
\end{table*}%

\section{Required Injection Parameters}
\label{appendix:parameters}

This section compares the number of required parameters in the two alternative injection methods discussed in section \ref{sec:architecture}: a multihead attention injection mechanism which is based on previous methods for combining external knowledge with BERT  and a novel lightweight gated injection mechanism. 

\paragraph{Multihead attention injection} 
In multihead attention injection (equations 
\ref{eq_multiheadattention_injection_1} to \ref{eq_multiheadattention_injection_2}), the keys are provided by BERT’s representation from the injection layer $\mathbf{H^i}$ and the queries are the injected information $\mathbf{I}$.
Multihead attention requires the following weight matrices $\mathbf{W}$ and biases $\mathbf{b}$ to transform queries, keys and values (indicated by $Q$, $K$ and $V$) and transform the attention output (indicated by $O$): 
\begin{align}
\begin{split}
\text{params}(\text{MultiHeadAttentionInjection})=&
\text{params}(\mathbf{W^K},\mathbf{W^Q},\mathbf{W^V},\mathbf{b^K},\mathbf{b^Q},\mathbf{b^V})\\
 &+ \text{params}(\mathbf{W^O},\mathbf{b^O})\\
 =& D(2D+ 2E + 4 D)\\
 \mathbf{W^Q}, \mathbf{W^O} \in \mathbb{R}^{D \times D},\mathbf{W^K},\mathbf{W^V}& \in \mathbb{R}^{E \times D},\mathbf{b^K},\mathbf{b^Q},\mathbf{b^V},\mathbf{b^O} \in \mathbb{R}^{D}
 \end{split}
\end{align}
where $D$ indicates BERT's hidden dimension and $E$ indicates the dimensionality of the injected embeddings. When injecting  embeddings with $D=300$ (see section \ref{sec:injected_embd}) 
into BERT$_\text{BASE}$ with $E=768$, this amounts to $\approx 1.6M$ new parameters.

\paragraph{Gated injection}
The proposed gated injection method (equations 
\ref{eq_gating_projection} to \ref{eq_gating})
only introduces the weights and biases from the projection layer, as well as the gating vector:
\begin{align}
\begin{split}
\text{params}(\text{GatedInjection})=& \text{params}(\mathbf{W^P},\mathbf{b^P},\mathbf{g})\\
 = &D(E+ 2).\\
 \mathbf{W^P} \in \mathbb{R}^{D \times E},&\mathbf{b^P} \in \mathbb{R}^D,\mathbf{g} \in \mathbb{R}^D
  \end{split}
\end{align}
Therefore, injecting  embeddings  with $D=300$ into BERT$_\text{BASE}$  requires  $\approx 0.2M$ new parameters. Our proposed gated injection mechanism only requires 14\% of the parameters used in a  multihead attention injection mechanism.
Using fewer parameters  results in a smaller model which is especially beneficial for injecting information during finetuning, where small learning rates and few epochs make it difficult to learn large amounts of new parameters.

\newpage

\section{Best Hyper-Parameters}
\label{appendix:hyperparams}

Hyper-parameters were chosen based on development set F1 scores.

\begin{table}[h!]
  \small
  \centering
    \begin{tabular}{lrrrrr}
    \toprule
{} &  MSRP &  Quora &  \multicolumn{3}{c}{SemEval}  \\ 
  &  &  &  A & B & C \\
    	\midrule
batch size & 32 & 32 & 16 & 32 & 16 \\
\midrule
\multicolumn{6}{l}{BERT}\\
Learning rate (1st seed) & 5e-5 &  2e-5  &    3e-5  &  2e-5   &    2e-5  \\ 
Learning rate (2nd seed) & 5e-5 &  2e-5  &    2e-5  &  2e-5   &    3e-5  \\ 
\midrule
\multicolumn{6}{l}{AiBERT with dependency-based embeddings}\\
Learning rate (1st seed) &3e-5  & 3e-5  &  2e-5    &   3e-5 &  2e-5   \\ 
Learning rate (2nd seed) & 5e-5 & 2e-5   &    2e-5  &  5e-5  &   2e-5  \\ 
\midrule
\multicolumn{6}{l}{AiBERT with counter-fitted embeddings}\\
Learning rate (1st seed) &5e-5  & 2e-5  &    2e-5  & 3e-5   &  2e-5   \\ 
Learning rate (2nd seed) & 5e-5 &  3e-5  &    5e-5  & 3e-5   &  2e-5   \\ 
\midrule
\multicolumn{6}{l}{GiBERT with dependency-based embeddings}\\
Learning rate (1st seed) & 2e-5 &  3e-5  &    2e-5  &  3e-5   &    2e-5  \\ 
Learning rate (2nd seed) & 3e-5 &  2e-5  &    2e-5  &  5e-5   &    3e-5  \\ 
\midrule
\multicolumn{6}{l}{GiBERT with counter-fitted embeddings}\\
Learning rate (1st seed) & 5e-5 &  2e-5  &    2e-5  &  5e-5   &    2e-5  \\ 
Learning rate (2nd seed) & 5e-5 &  3e-5  &    3e-5  &  5e-5   &    3e-5  \\ 
    \bottomrule
    \end{tabular}%
   \caption[Tuned hyper-parameters for BERT-based models.]{Tuned hyper-parameters for BERT-based models.}
       \label{tab:hyper_params_GiBERT}
\end{table}%

\section{Gating Parameter Analysis}
\label{sec:gating_vector_analysis}

\begin{figure*}[b!]
  \centering
  \includegraphics[height=6.5cm]{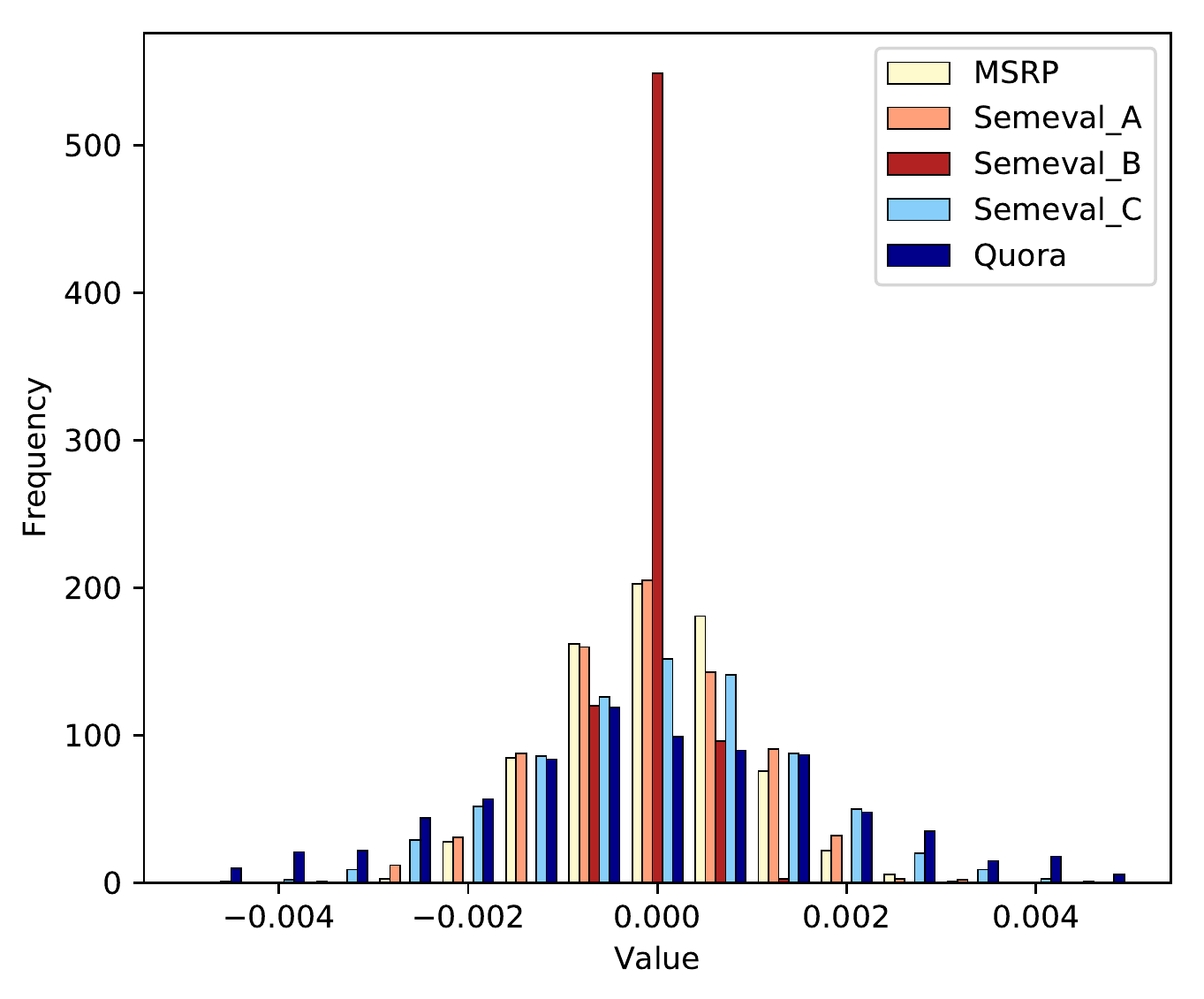}
    \includegraphics[height=6.5cm]{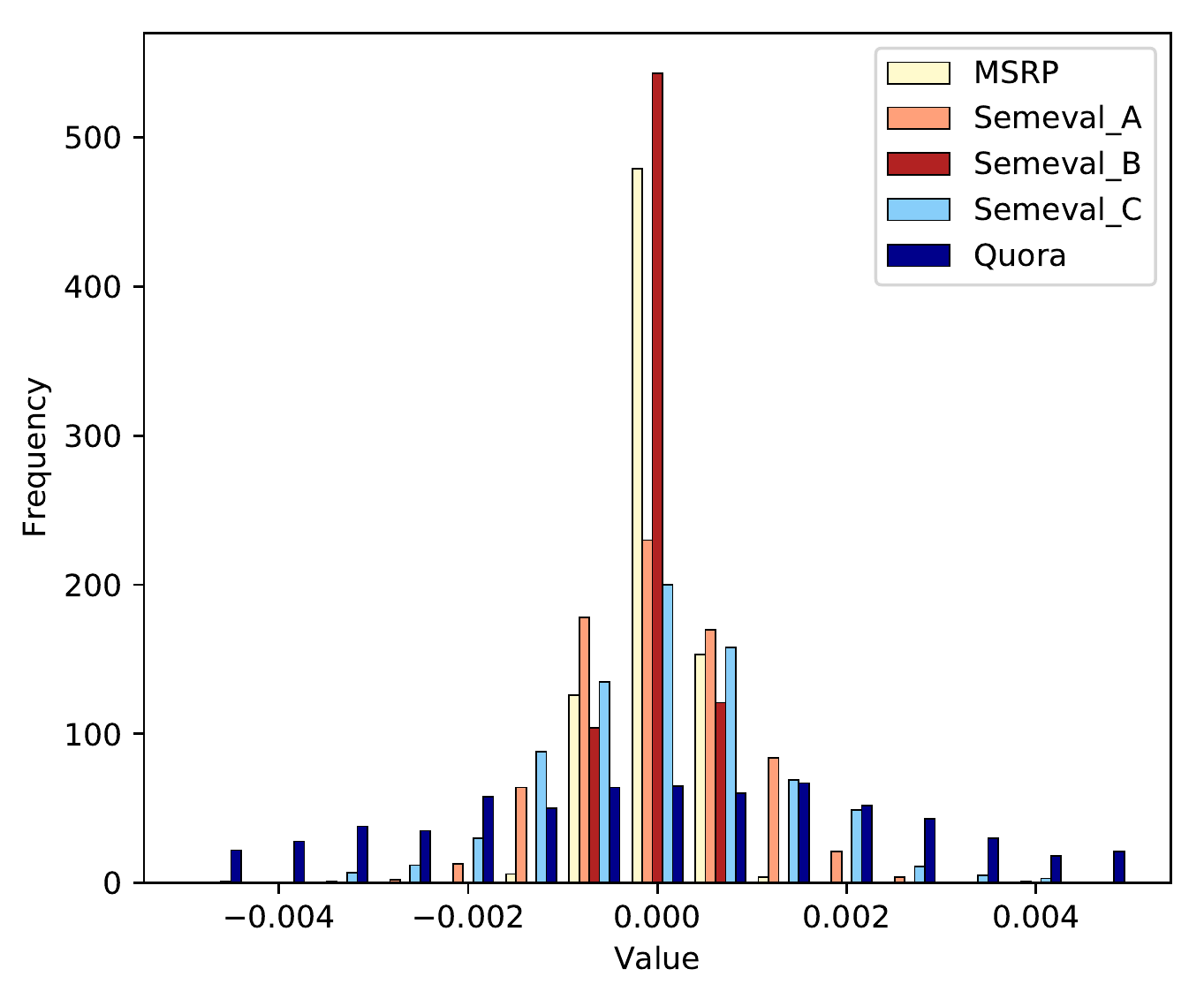}
   \caption{Histogram of the 768-dimensional gating vector $g$ across datasets for GiBERT with counter-fitted embeddings (left) and GiBERT with dependency embeddings (right). 
} 
  \label{fig_gating_vector}
\end{figure*}

As described in section 
\ref{sec:model}, the gating parameters $g$ in our proposed model are initialised as a vector of zeros. During training, the model can learn to gradually inject external information by adjusting gating parameters to $>0$ for adding, or $<0$ for subtracting injected information along certain dimensions. Alternatively, injection stays turned off if all parameters remain at zero. Figure~\ref{fig_gating_vector} shows a histogram of learned gating vectors for our best GiBERT models with counter-fitted (left) and dependency embedding injection (right). 
On most datasets, the majority of parameters have been updated to small non-zero values, letting through controlled amounts of injected information without completely overwriting BERT's internal representation.
Only on Semeval B (with 4K instances the smallest of the datasets, compare section~
\ref{sec:tasks}), more than 500 of the 768 dimensions of the injected information stay blocked out for both model variants. The gating parameters also filter out many dimensions of the dependency-based embeddings on MSRP (the second smallest dataset). 
This suggests that models trained on smaller datasets may benefit from slightly longer finetuning or a different gating parameter initialisation to make full use of the injected information.\footnote{Note that we train models for the same number of epochs, but one epoch uses all training examples contained in the dataset. This gives models trained on larger datasets more opportunity to update their parameters.} 

\end{document}